\pdfoutput=1

\documentclass[11pt]{article}

\usepackage{tikz}
\usepackage{mathtools}
\usepackage{multirow}
\usepackage{graphicx}
\usepackage{amsmath}
\usepackage{stfloats}
\usepackage[]{acl}

\usepackage{times}
\usepackage{latexsym}

\usepackage[T1]{fontenc}

\usepackage[utf8]{inputenc}

\usepackage{microtype}

\usepackage{inconsolata}

\title{Length-Aware Multi-Kernel Transformer for Long Document Classification}

\author{
    Guangzeng Han$^\dagger$ \quad 
    Jack Tsao$^\ddagger$ \quad 
    Xiaolei Huang$^\dagger$\thanks{\,\, Corresponding author} \\
    $^\dagger$ Department of Computer Science, University of Memphis, United States \\
    $^\ddagger$ Department of Neurology, Langone School of Medicine, New York University, United States \\
    \texttt{ghan@memphis.edu, jack.tsao@nyulangone.org, xiaolei.huang@memphis.edu}
}

\begin{document}
\maketitle
\begin{abstract}
Lengthy documents pose a unique challenge to neural language models due to substantial memory consumption. 
While existing state-of-the-art (SOTA) models segment long texts into equal-length snippets (e.g., 128 tokens per snippet) or deploy sparse attention networks, these methods have new challenges of context fragmentation and generalizability due to sentence boundaries and varying text lengths.
For example, our empirical analysis has shown that SOTA models consistently overfit one set of lengthy documents (e.g., 2000 tokens) while performing worse on texts with other lengths (e.g., 1000 or 4000).
In this study, we propose a \textbf{L}ength-\textbf{A}ware \textbf{M}ulti-\textbf{K}ernel \textbf{T}ransformer (\textit{LAMKIT}) to address the new challenges for the long document classification.
LAMKIT encodes lengthy documents by diverse transformer-based kernels for bridging context boundaries and vectorizes text length by the kernels to promote model robustness over varying document lengths.
Experiments on five standard benchmarks from health and law domains show LAMKIT outperforms SOTA models up to an absolute 10.9\% improvement.
We conduct extensive ablation analyses to examine model robustness and effectiveness over varying document lengths.
\footnote{Code available at~\url{https://github.com/trust-nlp/LAMKIT}}
\end{abstract}

\section{Introduction}

Lengthy documents widely exist in many fields, while the input limit  of transformer models prevents developing powerful pre-trained language models on those long documents, such as BERT~\cite{devlin2019bert} and RoBERTa~\cite{liu2019roberta}.
For example, a recent study shows that clinical documents have grown over 60\% longer in a decade~\cite{rule2021length}.
Truncation is a common strategy to handle long documents and fit the input limit of BERT-based classifiers, however, the method may lose many critical contexts beyond the first 512 tokens and hurdle model effectiveness.
Auto-regressive large language models (LLMs), such as ChatGPT~\cite{openai2022chatgpt} show their great ability at processing long documents, however the training object of these LLMs is to prediction the next token, which is inconsistent with the text classification task.
In other words, supervised fine-tuning on these domain specific data may not improve the performance of LLMs on these classification task.
Therefore, researchers focus on prompting methods~\cite{Wei2022Chain,chen2023need,song23ZeroPrompt,sun2024Contextual,zhang2024unlocking} or decoding strategies~\cite{wang2023selfconsistency} rather than fine-tuning~\cite{xiong2024large} to help the LLMs categorize text better.
This reality makes LLMs limited in this scenario.
Compare with these methods, developing discriminative transformer models that can model long documents is a more direct and effective solution to handle the long document classification task.

Among existing transformer-based models, long document modeling has two major directions, hierarchical transformer and sparse attention~\cite{dong2023survey, qin2023nlp}.
The hierarchical approach~\cite{wu2021hi, chalkidis2022lexglue, dai2022revisiting, li2023hipool, chalkidis2023lexfiles} splits document into small text chunks (e.g., 128 tokens) so that long document models can take shorter input per step.
As the self-attention in transformer-style models causes quadratic complexity \( O(n^2) \), the sparse attention aims to lower the complexity to linear and reduce context fragmentation caused by the segments~\cite{beltagy2020longformer, zaheer2020big, guo2022longt5, zhang2023adaptive}.
For example, sparse attention in Longformer~\cite{beltagy2020longformer} lifts up the input limit from 512 tokens to 4096 tokens.
Popular evaluation benchmarks also switch from social media data (e.g., IMDb and Amazon reviews~\cite{wu2021hi})
to more complex data in health and legal domains~\cite{qin2023nlp, chalkidis2022lexglue}.
For example, the median document length of IMDb is only 225 tokens~\cite{li2023hipool}, which is much smaller than the lengths in Table~\ref{tab:data}.
Indeed, document lengths vary across datasets, and model performance can vary across length-varied corpora~\cite{li2023hipool}.
However, very few studies have examined if long document models can handle varying-length texts, ranging from short to extremely long.
A common question is: \textit{will a long document model be capable to maintain robust performance across varying-length data?}
Our analysis on SOTA baselines in Figure~\ref{fig:Quarter} says ``No.''

\begin{table*}[htp]
\centering
\begin{tabular}{c||cccccc|ccc}
\multirow{2}{*}{Dataset} & \multicolumn{3}{c}{Length-Quantile} & \multirow{2}{*}{L-mean} & \multirow{2}{*}{Size} & \multirow{2}{*}{$|$Label$|$} & \multicolumn{3}{c}{Splits}\\
 & 25\% & 50\% & 75\% &  &  &  & Train & Valid & Test\\ \hline
Diabetes & 408 & 608 & 945 & 720 & 1,265 & 10 & 885 & 190 & 190\\ 
MIMIC & 1,432 & 2,022 & 2,741 & 2,200 & 11,368 & 50 & 8,066 & 1,753 & 1,729\\
ECtHR A/B & 668 & 1,328 & 2,627 & 2,139 & 11,000 & 11 & 9,000 & 1,000 & 1,000\\
SCOTUS & 3,723 & 7,673 & 12,275 & 9,840 & 7,800 & 14 & 5,000 & 1,400 & 1,400\\
\end{tabular}
\caption{Statistics of average token count per document (L-mean), data size (Size), and unique labels ($|$Label$|$).}
\label{tab:data}
\end{table*}

To understand the length effects and encounter the long document challenges, we conduct extensive analysis and propose \textbf{L}ength-\textbf{A}ware \textbf{M}ulti-\textbf{K}ernel \textbf{T}ransformer (\textit{LAMKIT}) for robust long document classification. 
LAMKIT diversifies learning processes by a multi-kernel encoding (MK) so that the model can capture contexts from different perspectives.
The MK contains multiple neural encoders with diverse kernel sizes and can relieve context fragmentation caused by a unique segment encoder on short text chunks.
LAMKIT promotes model robustness over varying-length documents by a length-aware vectorization (LaV) module. 
The LaV encodes length information in a hierarchical way, position embedding on segment and length vectors on document level.
We compare LAMKIT with 8 domain-specific models on five datasets (MIMIC-III~\cite{johnson2016mimic}, SCOTUS~\cite{chalkidis2022lexglue}, ECtHR-A~\cite{chalkidis2019neural} and ECtHR-B~\cite{chalkidis2021paragraph}, Diabetes~\cite{stubbs2019cohort}) from health and legal domains evaluated by F1 and AUC metrics.
Additionally, we also conduct a case study on the performance of ChatGPT in these tasks.
Classification results demonstrate that our LAMKIT approach's outperforms competitive baselines by an absolute improvement of up to 10.9\%.
We conduct further experiments on the length-varying effects and ablation analysis to examine the effectiveness of our individual modules.

\section{Data}
\label{sec:data}

We have retrieved five publicly available dataset, Diabetes~\cite{stubbs2019cohort}, MIMIC-III~\cite{johnson2016mimic}, ECtHR-A~\cite{chalkidis2019neural}, ECtHR-B~\cite{chalkidis2021paragraph}and SCOTUS~\cite{chalkidis2022lexglue}, which are popular benchmarks for the long document classification.
We obtained \textit{Diabetes}~\cite{stubbs2019cohort} from the 2018 National NLP Clinical Challenges (n2c2) shared task with a collection of longitudinal patient records and 13 selection criteria annotations.
We exclude 3 annotations due to less than 0.5 inter-rater agreements and discard documents with fewer than 40 tokens. 
\textit{MIMIC-III} (Medical Information Mart for Intensive Care)~\cite{johnson2016mimic} is a relational database that contains patients admitted to the Intensive Care Unit (ICU) at the Beth Israel Deaconess Medical Center from 2001 to 2012. 
We follow previous work~\cite{mullenbach2018explainable, vu2021label} to select discharge summaries and use the top 50 frequent labels of International Classification of Disease codes (9th Edition, ICD-9), which are types of procedures and diagnoses during patient stay in the ICU.
\textit{ECtHR-A} collects facts and articles from law case descriptions from the European Court of Human Rights’ public database~\cite{chalkidis2019neural}.
Each case is mapped to the articles it was found to have violated in the ECHR, while in \textit{ECtHR-B}~\cite{chalkidis2021paragraph}, cases are mapped to a set of allegedly violated articles.
\textit{SCOTUS} is a data collection of US Supreme Court (the highest US federal court) opinions and the US Supreme Court Database (SCDB)~\cite{spaeth2020Supreme} with cases from 1946 to 2020.
SCOTUS has 14 issue areas, such as Criminal Procedure, Civil Rights, and Economic Activity.
We summarize data statistics and splits in Table~\ref{tab:data}.

Table~\ref{tab:data} shows each data has a varying length range, a critical yet under-explored question is: does the varying length effect model performance or will models be generalizable across all lengths? 
For example, the document length in Table~\ref{tab:data} is either less than a few hundred or over ten thousand tokens surpassing input limitations of regular transformer-style models (e.g., BERT), and there are significant length variations across the data.
While studies~\cite{dong2023survey} have achieved improving performance overall to encode more contexts beyond the 512 token limit, there is very few work examining the effects of varying document lengths over model robustness.
To answer the question, we conduct an exploratory analysis of existing state-of-the-art (SOTA) models and evaluate their performance.

\begin{figure}[htp]
\centering
\includegraphics[width=0.5\textwidth]{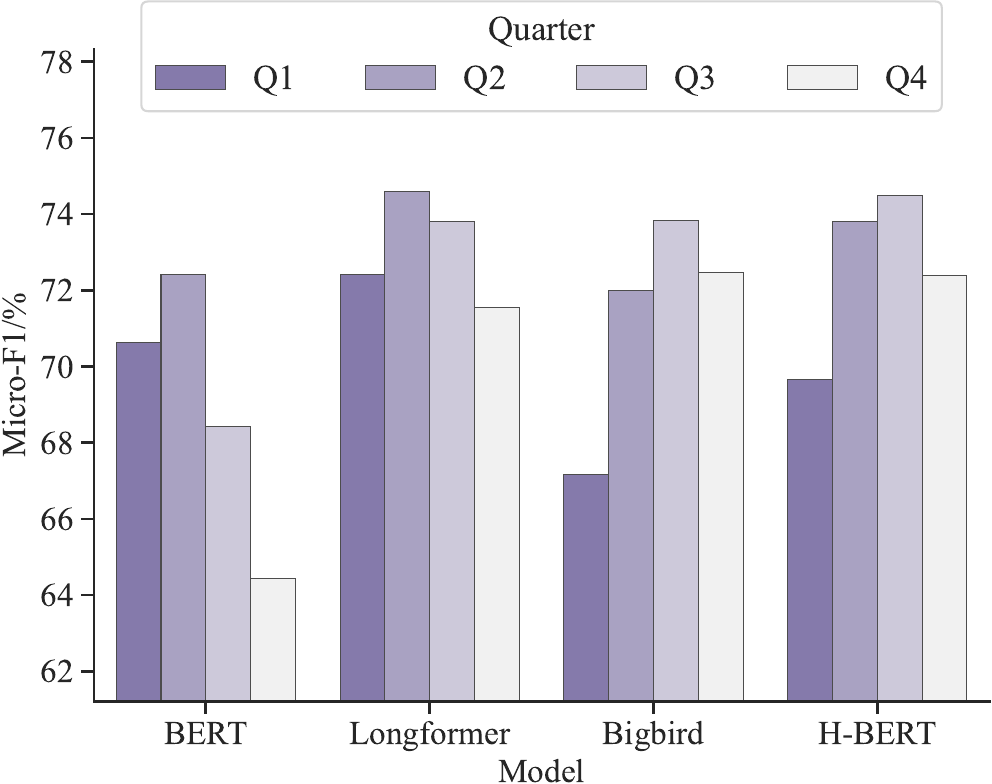}  
\caption{Average performance on quarter splits for four state-of-the-art baselines. The length boundaries of quarters are shown in Table~\ref{tab:data}. Detailed performance scores are presented in Table~\ref{tab:varying} }
\label{fig:Quarter}
\end{figure}

Our exploratory analysis follows existing studies~\cite{mullenbach2018explainable, dai2022revisiting, chalkidis2022lexglue, qin2023nlp} to split data, includes three state-of-the-art transformer classifiers (BigBird, Longformer, and Hierarchical BERT (H-BERT)) for long document and a BERT classifier, and evaluates models performance by F1-micro~(F1-$\mu$) score.
We refer to the details of experimental settings and SOTA baselines under the Experiments section.
For each quarter, we maintain similar data sizes and run the classifier multiple times to take average performance scores. 
Finally, we visualize the relation between model performance and document lengths in Figure~\ref{fig:Quarter}.

\begin{figure*}[htp]
\centering
\includegraphics[width=.92\textwidth]{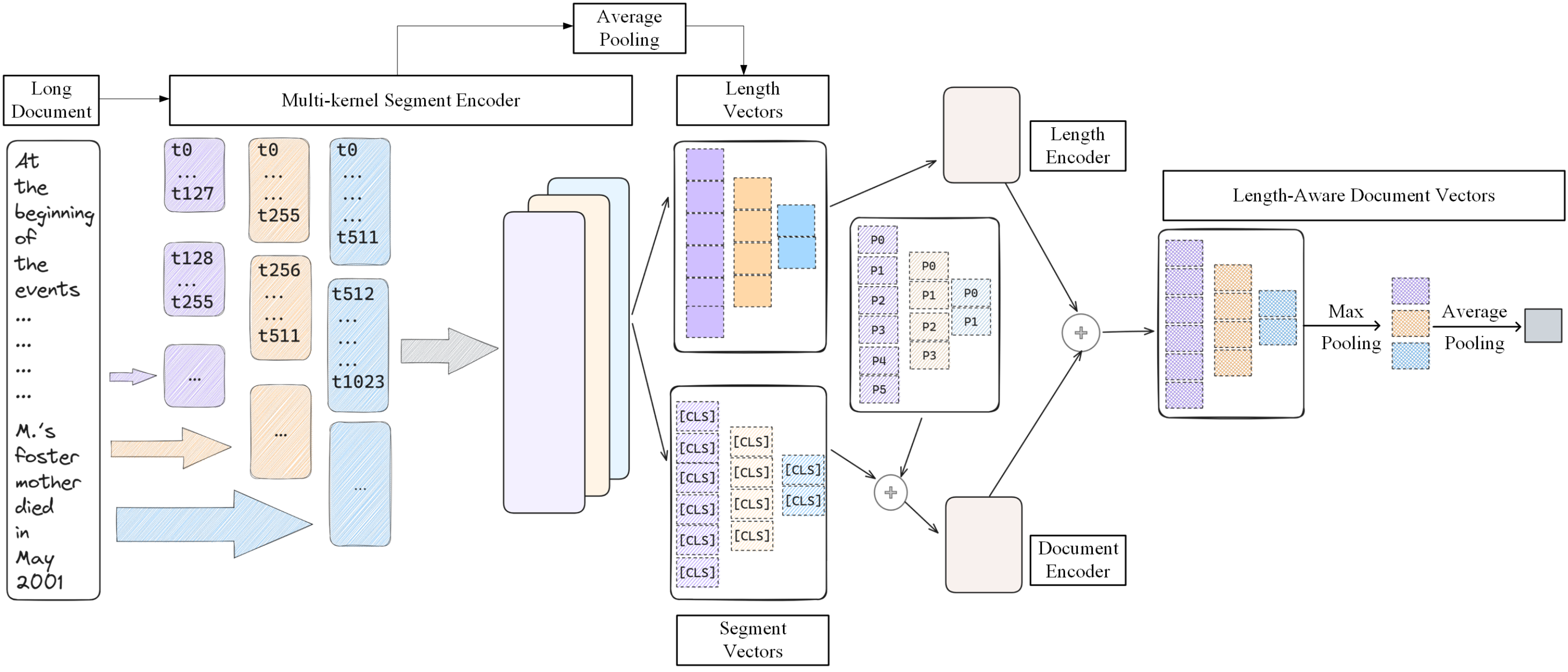}  
\caption{LAMKIT diagram overview. Our approach consists of three main components: multi-kernel encoding, length-aware vectorization, and hierarchical integration. We denote one color of segments and vectors per kernel. The arrows indicate model workflows, $\bigoplus$ is a sum operation.}
\label{fig:model}
\end{figure*}

Figure~\ref{fig:Quarter} shows that model performance varies across document lengths, posing a unique challenge to build robust models on varying lengthy data.
For example, while the SOTA classifiers achieve better scores on mid-lengthy texts, the performance drops significantly in either short (e.g., 400 tokens) or super long (e.g., 10K tokens) documents. 
The consistent observations can suggest that: 1) varying length can be a critical factor to make models perform better; 2) length-based splits are important to understand the capacity of classifiers on long documents.
The findings inspire us to propose the \textbf{L}ength-\textbf{A}ware \textbf{M}ulti-\textbf{K}ernel \textbf{T}ransformer (\textit{LAMKIT}) to encounter the length factor.

\subsection{Ethic and Privacy Concern}
 
All data used in this research is publicly accessible and has been stripped of identifying information. 
Our investigation is centered on computational techniques, and we do not gather data directly from individuals. 
Our institution's review board has confirmed that this research does not mandate an IRB approval.

\section{Length-Aware Multi-Kernel Transformer}

This section presents our Length-Aware Multi-Kernel Transformer (\textit{LAMKIT}) for robust long document classification in Figure~\ref{fig:model}.
LAMKIT consists of three major modules, 1) multi-kernel encoding, 2) length-aware vectorization, and 3) hierarchical integration, aiming to solve context fragmentation and augment model robustness on lengthy documents.
We deploy different encoding kernels to diversify text segments with various contexts.
Incorporating length as vectors can adapt classifiers across varying-length documents.
Finally, we elaborate on how to learn robust document representations via a hierarchical integration.

\subsection{Multi-kernel Encoding}

Multi-kernel Encoding (MK) aims to diversify context to segment and encode documents from multiple perspectives.
The mechanism is to solve the challenge of existing long document modeling methods~\cite{beltagy2020longformer, wu2021hi, dai2022revisiting, dong2023survey} –– splitting and vectorizing each document by a fixed size, which has been analyzed in our previous data section.
Our MK mechanism gets inspirations from TextCNN~\cite{kim2014convolutional}, which uses kernels of different sizes to convolve text representations. 
In contrast, our MK mechanism encodes each document into various sizes of text segments to obtain various feature representations. 
By learning diverse document features with varying-size text chunks, we can enrich representations of lengthy documents with various sizes.

Specifically, we empirically choose a set of kernel sizes (e.g. $m = \{128, 256, 512\}$ for the MIMIC dataset) to split and vectorize the long documents.
Following the CNN, we tried the stride ranging between $(2/3*m,m)$, but we did not get significant improvements. 
Therefore the stride of all kernels is set to its kernel size such that two adjacent segments do not overlap.
In the later section, our ablation analysis shows that the major performance drops come from the number of kernels.
We infer the performance of kernel and stride sizes as encoding contexts with different kernels is more critical to augment classifiers on lengthy documents.
For each chunk size of text, we deploy a pre-trained RoBERTa model~\cite{liu2019roberta} so that our MK has enriched representations for the varied text chunks.
While our MK mechanism allows other Transformer variants, we choose the RoBERTa to keep consistent with existing SOTA approaches~\cite{chalkidis2022lexglue, li2023comparative, dong2023survey} for fair comparisons.
We take the embedding of the ``[CLS]'' token from each text chunk to represent its segment vector and feed to the following operation, combining with the segment position embedding of length-aware vectorization.

\subsection{Length-aware Vectorization}

We propose the Length-aware Vectorization (\textit{LaV}) to incorporate lengthy contexts and augment model generalizability, as our Figure~\ref{fig:Quarter} presents that the model performance varies across document lengths.
LaV achieves the grand goal by two levels: text chunk and document. 
On the text chunk level, we encode length information by the segment position embedding, and on the document level, we vectorize text length with MK outputs. 

\paragraph{Segment Position Embedding} vectorizes positions of text chunks into a learnable embedding by a Transformer encoder in Equation~\ref{eq:pe}, where $|d|$ refers to the embedding size, $i$ is the column index of a vector scalar, and $pos$ is the index of the text chunk.
For example, if we segment a 1024-token document into 15 chunks (with a stride) by the 128 kernel encoder, the total will be the 15 and the second chunk's index (pos) will be 2.
Similarly, we can obtain segment position embeddings for other multi-kernel encoders and equip the segment vectors from the MK step with the length information, segment position.
Finally, we sum the segment position embeddings up with the segment vectors and feed them to the document encoder.

\begin{equation}
\text{\resizebox{0.167\linewidth}{!}{$PE_{(pos, i)}$}} = 
\begin{cases}
\sin \left( \frac{pos}{10000^{2i/|d|}} \right), & \text{if } i \text{ is even}\\
\cos \left( \frac{pos}{10000^{2i/|d|}} \right), & \text{if } i \text{ is odd}
\end{cases}
\label{eq:pe}
\end{equation}

Note that, our position embedding \textbf{differs} from previous studies.
For example, majority of long document classifiers~\cite{wu2021hi, li2023recurrent, zhang2023adaptive} deploy position embeddings for tokens rather than the segment.
There is one close study~\cite{dai2022revisiting} that utilizes segment position embedding in classification models.
In contrast, our position embedding diversifies segment positions from multiple kernels, aiming to incorporate text lengths and augment model generalizability over varying text lengths.

\paragraph{Length Vectors} encode document length information into feature vectors.
Instead of directly encoding a length scalar into a vector, we obtain the length vectors by applying averaging pooling over each MK encoder's outputs and vectorizing the chunk sizes per document by the position embedding.
The length vectors not only encode document lengths by chunk sizes but also implicitly incorporate lengthy contexts from the MK encoders.
Finally, we feed the length vectors into the length encoder to obtain learnable length-aware vectors, which will be integrated with the document encoder's outputs.

\subsection{Hierarchical Integration}

We obtain length-aware document representations through the hierarchical integration process from segment and length vectors.
The integration process starts with a document encoder to encode segment vectors and a length encoder to encode length vectors.
Both modules are Transformer~\cite{vaswani2017attention} encoders but serve different purposes –– while both encoders take length-related vectors, the document encoder focuses on learning diversified contexts from the MK encoders and the length encoder focuses on incorporating varying length features.
We then combine the two encoders' outputs by a sum operation and feed the integration to a hierarchical pooling process to obtain length-aware document vectors.

\paragraph{Hierarchical pooling operations} has two major processes in order, max pooling and average pooling. 
The max pooling aims to squeeze length-aware multidimensional representations of text chunks from the length and document encoders.
We concatenate the pooling outputs and feed them to the average pooling operation.
The average pooling aggregates the length-aware segment features into the length-aware document vectors.
Finally we feed the document vectors to linear layer for classification.
Our tasks cover both binary and multi-label classifications.
We deploy a sigmoid function for binary prediction and a softmax function for the multi-label task.

\section{Experiments}
We follow the previous studies~\cite{mullenbach2018explainable, stubbs2019cohort, chalkidis2022lexglue} on lengthy document to preprocess data and split data into training, validation, and test, as in Table~\ref{tab:data}.
We follow SOTA baselines to set up our evaluation experiments.
Our results include F1 and AUC metrics, covering both micro ($\mu$) and macro (m) variations.

Our evaluation presents performance comparisons and ablation analysis to understand the length effects and the models better.
More details of the hyperparameter settings for the baselines and LAMKIT are in the Appendix~\ref{sec:implementation}, which allows for experiment replications.

\begin{table*}[ht]
\centering
\resizebox{\textwidth}{!}{
\begin{tabular}{l||cccc|cccc|cccc|cccc|cccc}
\multirow{2}{*}{Model}& \multicolumn{4}{c|}{Diabetes} & \multicolumn{4}{c|}{MIMIC} & \multicolumn{4}{c|}{ECtHR-A} & \multicolumn{4}{c|}{ECtHR-B}& \multicolumn{4}{c}{SCOTUS} \\
\multicolumn{1}{c||}{}  & F1-$\mu$ & F1-m      & AUC-$\mu$ & AUC-m     & F1-$\mu$ & F1-m    & AUC-$\mu$ & AUC-m   & F1-$\mu$ & F1-m    & AUC-$\mu$ & AUC-m  & F1-$\mu$ & F1-m     & AUC-$\mu$ & AUC-m   & F1-$\mu$ & F1-m    & AUC-$\mu$ & AUC-m   \\ \hline
 $ \text{BERT}_{\textit{First}} $           & \underline{72.0}  & 43.2  & 86.9  & 72.4  & 56.8  & 47.0 & 87.1 & 84.0 & 64.2  & 52.6 & 91.6 & 88.6 &73.3 & 67.6 & 93.1 & 91.4& 73.9  & 61.6 & \underline{95.9} & 90.0 \\
$ \text{BERT}_{\textit{Last}}  $            & 68.7  & 39.1  & 87.2  & 72.2  & 51.3  & 41.5 & 84.8 & 81.4 & 66.1  & 59.1 & 93.7 & 91.3 &75.1 & 65.7 & 94.5 & 93.0& 66.9  & 53.1 & 93.6 & 87.2 \\
Longformer                                  & 71.5  & 41.2  & \underline{88.4}  & 71.6  & \underline{67.2}  & 58.2 & 92.5 & 89.8 & \underline{71.4}  & 59.0 & 95.4 & 93.3 &\underline{79.6} & \underline{73.1} & 95.2 & 94.0& 74.3  & 62.9 & 95.6 & 89.9 \\
BigBird                                     & 71.9  & 42.5  & \textbf{88.5}  & \textbf{76.4}  & 65.3  & 56.8 & 92.3 & 89.7 & 70.2  & \underline{61.8} & 93.8 & 91.8 &78.9 & 70.3 & \underline{95.5} & 93.8& 72.3  & 60.6 & 94.3 & 89.7 \\
H-BERT                                      & 70.4  & \underline{46.0}  & 83.2  & 69.7  & 66.9  & \underline{60.6} & \underline{92.6} & \underline{90.2} & 70.4  & 57.7 & \underline{95.7} & \underline{93.9} &79.2 & 72.0 & 95.4 & \underline{94.4}& \underline{76.6}  & \textbf{68.0} & 95.5 & \textbf{95.0} \\ \hline
LAMKIT                                      & \textbf{73.4}  & \textbf{49.9}  & \underline{88.4}  & \underline{74.5}  & \textbf{69.5}  & \textbf{63.7} & \textbf{93.3} & \textbf{91.2} & \textbf{73.0}  & \textbf{65.0} & \textbf{96.0} & \textbf{94.7} &\textbf{80.2} & \textbf{74.4} & \textbf{95.8} & \textbf{94.7}& \textbf{78.5}  & \underline{67.8} & \textbf{97.1} & \underline{94.9} \\
 $\overline{\Delta}$                        & $\uparrow$2.5   & $\uparrow$6.9   & $\uparrow$1.6   &$\uparrow$2.0   & $\uparrow$8.0   & $\uparrow$10.9 &$\uparrow$3.4  & $\uparrow$4.2  & $\uparrow$4.5   & $\uparrow$7.0  & $\uparrow$2.0  & $\uparrow$2.9  &$\uparrow$3.0 & $\uparrow$4.7 & $\uparrow$1.1 & $\uparrow$1.4& $\uparrow$5.7   & $\uparrow$6.6  & $\uparrow$2.1  & $\uparrow$4.5 
\end{tabular}}

\caption{Overall performance in percentages of F1 and AUC metrics, both micro ($\mu$) and macro (m). We \textbf{bolden} the best performance and \underline{underline} the second best value. $\overline{\Delta}$ denotes the absolute improvement of LAMKIT over the baselines average.}
\label{tab:all_results}
\end{table*}

\subsection{Baselines}

To demonstrate the effectiveness of LAMKIT, we compare it against both hierarchical transformer and sparse attention transformer SOTA baselines for long-document modeling, as well as with regular BERT.
Although our LAMKIT has no theoretical length limit, we set the text length to 4096 for all experiments for a fair comparison, except for BERT which is 512.

Our experiments utilize baseline hyperparameters that achieved their best results in the previous studies.
For example, we take publicly released models or source codes to train long document classifiers.
As our data come from health and legal domains, we choose the pre-trained models on the domain data.
For example, we report the performance of Clinical-Longformer~\cite{li2023comparative} and Legal-Longformer~\cite{chalkidis2023lexfiles} on health and legal data, respectively, instead of the vanilla Longformer~\cite{beltagy2020longformer}.

\paragraph{BERT} includes classifiers built on domain-specific pre-trained BERT models. 
Specifically, we include two types of pre-trained BERT model, \textit{Legal-BERT}~\cite{chalkidis2020legal} for the legal data and \textit{RoBERTa-PM-M3}~\cite{lewis2020pretrained} for the clinical data, which achieved the best performance on broad text classification tasks in legal and clinical domains.
Due to the input limit, the BERT baselines truncate and only take 512 tokens per entry.
We experiment two types of truncation, first and last 512 tokens of each data entry, and name the two types as $BERT_{First}$ and $BERT_{Last}$.

\paragraph{Hierarchical BERT} (\textit{H-BERT}) splits long document into equal-length segments, hierarchically integrate segment features into document vectors, and yield predictions on the document vectors~\cite{dai2022revisiting, qin2023nlp, dong2023survey}.
We follow the existing SOTA studies that achieved the best results using the H-BERT in health~\cite{dai2022revisiting} and legal~\cite{chalkidis2022lexglue} domains.
The H-BERT models are close to our hierarchical architecture, while the H-BERT models do not incorporate our proposed multi-kernel mechanism (MK) and length vectors.
If LAMKIT achieves better performance, the improvements over the H-BERT can prove the effectiveness of adapting varying-length texts.

\paragraph{Longformer} \cite{beltagy2020longformer} solves the 512-length limit by replacing self-attention with a local (sliding window) attention and unidirectional global attention and thus can process sequences up to 4096 tokens.
We deploy domain-specific Longformer to keep consistent experimental settings.
Specifically, we utilize \textit{Clinical-Longformer}~\cite{li2023comparative} and \textit{Legal-Longformer}~\cite{chalkidis2023lexfiles} to build our document classifiers for the health and legal data, respectively.

\paragraph{BigBird} deploys a block sparse attention to relieve the length limit that reduces the Transformer quadratic dependency to linear~\cite{zaheer2020big}. 
BigBird utilizes a fusion of local, global, and random attention, extending the maximum processable sequence length to 4096 tokens.
We utilize its domain-specific variants, Clinical-BigBird~\cite{li2023comparative} and Legal-Bigbird~\cite{dassi2021legal} to conduct experiments.

\section{Result Analysis}
This section reports the performance of SOTA baselines and LAMKIT in terms of F1 and AUC metrics, both micro ($\mu$) and macro (m) modes.
Besides the overall performance, we examine varying-length effects and conduct ablation analysis on our individual modules (e.g., MK and LaV). 
The results show that LAMKIT not only surpasses the baselines by a large margin on long documents from both health and legal domains but also shows more stable performance on documents of varying lengths.

\subsection{Overall Performance}
We present the results of long document classification benchmarks in Table~\ref{tab:all_results} that our LAMKIT significantly outperforms the other SOTA baselines.
For example, compared to the baselines' average performance, LAMKIT shows an improvement of 4.7\% in F1-micro and 7.2\% in F1-macro.
Long document models do not perform better than regular BERT models on shorter texts.
For example, $BERT_{first}$ outperforms most of the SOTA baselines on Diabetes, of which 50\% clinical notes are less than 608 tokens.
In contrast, we can observe our LAMKIT is robust on both shorter and longer text documents, highlighting the unique contribution and effectiveness of our approach.

Document characteristics of health and legal data can impact baselines performance.
For example, we find that H-BERT performs better on the SCOTUS compared to models with sparse attention networks (e.g., Longformer and BigBird), while its performance on other datasets is comparable.
We infer this as the SCOTUS dataset has clear segment boundaries that H-BERT can utilize the boundaries as segments, however, other data is compressed and dense, which can cause context fragmentation~\cite{beltagy2020longformer} and weaken effectiveness of H-BERT.
\textit{However}, our LAMKIT demonstrates superior performance on the issue, and we think the MK and length-aware vectors play critical roles, which is shown in our ablation analysis.

\subsection{Performance on Varying-length Splits}

To assess the model's robustness and generalizability across documents of varying lengths, we follow the approach described in the Data Section, dividing each dataset into quarters based on the lengths of the documents, ensuring similar data sizes in each quarter.

Table~\ref{tab:varying} presents F1-micro scores across four quarters of each dataset that LAMKIT outperforms baselines on most quarters across the datasets.
Surprisingly, SOTA baselines tend to favor and overfit one quarter data with a specific length, which does not exceed their input limit (e.g., 4096 for Longformer).
In contrast, our LAMKIT shows more generalizable performance across varying-length documents.
The stable performance of our LAMKIT highlights the effectiveness of our multi-kernel and length vectors in adapting classifiers on varying lengths and promoting classification robustness on the health and legal domains.

\begin{table*}[ht]
\centering
\resizebox{\textwidth}{!}{
\begin{tabular}{l||cccc|cccc|cccc|cccc|cccc}
\multirow{2}{*}{Model} & \multicolumn{4}{c|}{Diabetes} & \multicolumn{4}{c|}{MIMIC} & \multicolumn{4}{c|}{ECtHR-A} & \multicolumn{4}{c|}{ECtHR-B}& \multicolumn{4}{c}{SCOTUS} \\
 & Q-1 & Q-2 & Q-3 & Q-4 & Q-1 & Q-2 & Q-3 & Q-4 & Q-1 & Q-2 & Q-3 & Q-4 & Q-1 & Q-2 & Q-3 & Q-4 & Q-1 & Q-2 & Q-3 & Q-4 \\ \hline
$ \text{BERT}_{\textit{First}} $ & \underline{65.7} & \textbf{74.1} & 73.4 & 74.2 & 57.9 & 63.0 & 57.5 & 52.9 & 74.9 & 73.4 & 62.6 & 54.4 &79.6 & 77.3 & 70.7 & 70.7& \textbf{75.0} & 74.3 & 80.9 & 70.0 \\
$ \text{BERT}_{\textit{Last}} $ & 63.4 & 66.9 & 71.6 & 71.8 & 51.6 & 57.8 & 50.3 & 48.4 & 72.6 & 73.0 & 62.5 & 61.6 &77.8 & 79.5 & 73.4 & 73.0& 68.8 & 64.4 & 69.4 & 66.0 \\
Longformer & 64.6 & \underline{72.7} & 72.2 & 75.8 & \underline{63.8} & \underline{71.0} & \underline{68.1} & 66.4 & 79.0 & 74.0 & 72.4 & 65.7 &\textbf{84.4} & \textbf{81.9} & 79.4 & 76.4& 69.3 & 73.4 & 76.9 & 74.5 \\
BigBird & 61.0 & 72.1 & 71.7 & \textbf{79.9} & 62.9 & 70.2 & 66.3 & 62.6 & 68.8 & 65.9 & \underline{73.9} & \textbf{70.7} &77.8 & \underline{81.4} & \underline{80.1} & 77.0 & 65.3 & 70.4 & 77.2 & 72.1 \\
H-BERT & 61.2 & 67.6 & \underline{74.2} & 77.8 & 62.1 & 69.6 & 66.8 & \underline{66.5} & \underline{79.1} & \textbf{75.3} & 69.1 & 64.1 &\underline{81.7 }& 80.7 & 79.4 & \underline{77.1}& 64.2 & \underline{75.8} & \underline{82.9} & \underline{76.5} \\ \hline
LAMKIT & \textbf{66.0} & 71.2 & \textbf{77.0} & \underline{78.1} & \textbf{66.4} & \textbf{72.6} & \textbf{70.4} & \textbf{68.0} & \textbf{79.7} & \underline{74.6} & \textbf{74.3} & \underline{67.5} &79.4 & 80.8 & \textbf{80.3} & \textbf{80.0}& \underline{72.2} & \textbf{76.4} & \textbf{83.0} & \textbf{78.5} \\
$\overline{\Delta}$ & $\uparrow$2.8 & $\uparrow$0.5 & $\uparrow$4.4 & $\uparrow$2.2 & $\uparrow$6.7 & $\uparrow$6.3 & $\uparrow$8.6 & $\uparrow$8.6 & $\uparrow$4.8 & $\uparrow$2.3 & $\uparrow$6.2 & $\uparrow$4.2 &$\downarrow$-0.9 & $\uparrow$0.6 & $\uparrow$3.7 & $\uparrow$5.2& $\uparrow$3.7 & $\uparrow$4.7 & $\uparrow$5.5 & $\uparrow$6.7 \\
\end{tabular}
}
\caption{F1-micro scores across four quarters following our Figure~\ref{fig:Quarter}. We \textbf{bolden} the best performance and \underline{underline} the second best value. $\overline{\Delta}$ refers to the absolute improvement of LAMKIT over the average of baselines.}
\label{tab:varying}
\end{table*}

\subsection{Ablation Study}

We conduct an ablation analysis to assess the effectiveness of individual LAMKIT modules focusing on the multi-kernel mechanism (MK) and length-aware vectorization (LaV). 
Table~\ref{tab:ablation} shows the results of our anylysis.
\textit{w/o MK} replaces multi-kernel encoders with a single kernel encoder (RoBERTa) and shrinks segment vectors accordingly.
\textit{w/o LaV} removes length-related vectors and encoders from LAMKIT.
And, \textit{w/o MK and LaV} removes both MK mechanism and length-related encoding.

We can observe that removing one of the modules or removing all modules can significantly reduce model performance.
Replacing the MK mechanism can result in a 1.3\% and 1.9\% drop in F1-micro and F1-macro on average, respectively.
The performance drop indicates multi-kernel encoding mechanism can relieve context fragmentation to promote model performance by diversifying document representations.
Removing LaV leads to 1.3\% and 2.4\% drops in F1-micro and F1-macro on average, respectively.
The performance drop shows that the length information can be critical to building robust classifiers on the health and legal data.

We can observe the most significant performance drop in LAMKIT after removing both MK and LaV modules, with F1-micro and F1-macro scores decreasing by 2.8\% and 3.5\%, and AUC-micro and AUC-macro scores by 1.5\% and 1.8\%, respectively, demonstrating the effectiveness of these modules.

\section{Case Study on ChatGPT}

\begin{table*}[thp]
\centering
\resizebox{\textwidth}{!}{
\begin{tabular}{l|cccc|cccc|cccc|cccc|cccc}
\multirow{2}{*}{Model} & \multicolumn{4}{c|}{Diabetes} & \multicolumn{4}{c|}{MIMIC} & \multicolumn{4}{c|}{ECtHR-A}& \multicolumn{4}{c|}{ECtHR-B} & \multicolumn{4}{c}{SCOTUS} \\
                       & F1-$\mu$ & F1-m      & AUC-$\mu$ & AUC-m     & F1-$\mu$ & F1-m    & AUC-$\mu$ & AUC-m    & F1-$\mu$ & F1-m     & AUC-$\mu$ & AUC-m   & F1-$\mu$ & F1-m    & AUC-$\mu$ & AUC-m  & F1-$\mu$ & F1-m    & AUC-$\mu$ & AUC-m  \\ \hline
LAMKIT                    & 73.4  & 49.3  & 88.4  & 74.5  & 69.5  & 63.7 & 93.3 & 91.2 & 73.0  & 65.0 & 96.0 & 94.7 &80.2 & 74.4 & 95.8 & 94.7& 78.5  & 67.8 & 97.1 & 94.9 \\
w/o MK                & 72.1  & 47.6  & 88.2  & 72.3  & 68.5  & 61.9 & 92.8 & 90.5 & 72.0  & 62.7 & 95.5 & 93.9 &79.0 & 72.3 & 95.8 & 94.2& 76.7  & 66.3 & 97.0 & 93.3 \\
w/o LaV            & 71.5  & 42.1  & 87.5  & 72.7  & 68.4  & 62.9 & 93.0 & 90.8 & 71.5  & 64.2 & 95.6 & 94.3 &79.2 & 72.4 & 95.4 & 94.6& 77.6  & 66.6 & 97.1 & 93.1 \\
w/o MK and LaV                & 69.9  & 46.6  & 85.3  & 71.1  & 66.3  & 60.0 & 92.3 & 89.9 & 70.4  & 61.3 & 94.9 & 93.4& 78.0 & 70.7 & 94.1 & 93.4& 76.0  & 63.9 & 96.4 & 93.6
\end{tabular}}
\caption{Ablation performance of LAMKIT modules in F1 and AUC, both micro ($\mu$) and macro (m), shown in percentages.}
\label{tab:ablation}
\end{table*}

Large language models (LLMs) have achieved impressive performance on many generative tasks, such as long text summarization or long text QA. 
However, long text classification is a natural language understanding task, which makes fine-tuning the large model on such a task not a guaranteed improvement in classification accuracy.
Thus the dominant paradigms for text classification in LLMs are zero-shot learning and few-shot learning~\cite{lou2023prompt}.
To examine the ability of LLMs on the long document classification task, we utilize representative GPT-3.5-Turbo via \textit{ChatCompletion API}\footnote{\url{https://platform.openai.com/docs/guides/gpt/chat-completions-api}} in a zero-shot prompting strategy with multiple templated instructions summarized by \cite{lou2023prompt,chalkidis2023chatgpt,chen2023many}, and report the best performing template results. 
Due to privacy concerns and data usage agreement, we do not test ChatGPT~\cite{openai2022chatgpt} on MIMIC and Diabetes.
The results in Table~\ref{tab:gptf1} suggest that compared to our LAMKIT and also the chosen baseline models, ChatGPT still underperforms on long text classification tasks.
For the prompt template, we refer more details in the Appendix Figure~\ref{fig:template}.

\begin{table}[htp]
\centering
\resizebox{\columnwidth}{!}{
\begin{tabular}{l|cc|cc|cc}
\multirow{2}{*}{Model} & \multicolumn{2}{c|}{ECtHR-A} & \multicolumn{2}{c|}{ECtHR-B} & \multicolumn{2}{c}{SCOTUS} \\
                       & F1-$\mu$ & F1-m     & F1-$\mu$ & F1-m      & F1-$\mu$ & F1-m      \\ \hline
ChatGPT                & 51.1     & 47.7     & 54.0      & 60.8      & 49.9     & 42.0     
\end{tabular}
}
\caption{F1 metrics (in \%) of ChatGPT on Legal Data.}

\label{tab:gptf1}
\end{table}

\section{Related Work}

\subsection{Transformers for Text Classification}
Pretrained language models (PLMs) based on vanilla self-attention, such as BERT~\cite{devlin2019bert} and its variants~\cite{nerella2023transformers,he2021deberta,zhou2022eventbert,ma2021contributions, alsentzer2019publicly,jin2023understand}, have achieved state-of-the-art (SOTA) results in regular text classification tasks.
However, with their input typically limited to 512 tokens, truncation becomes necessary when handling long texts~\cite{ding2020cogltx}. 
Such truncation might cause the text to lose a significant amount of valuable information, thereby affecting the model's performance. 
Another option is to use the generative LLMs to categorize text, however, their architecture and training methods make them unsuitable for fine-tuning directly on text categorization tasks, thus previous studies have focused more on their zero-shot and few-shot performance\cite{han2024chainofinteraction,pan2024chainofaction,srivastava2023instance}.
Compare with these methods, long document modeling serves as a more directly solution to handle the long document classification task.

\subsection{Long Document Modeling}

To enable transformers to accept longer sequences, two primary approaches have been employed in long document modeling: efficient transformers (e.g., sparse attention transformers) and hierarchical transformers~\cite{dong2023survey}.
Hierarchical transformer models~\cite{li2023hipool,ruan2022histruct, chalkidis2023lexfiles} rely on chunking the text into slices of equal size and obtaining the document representation based on the representations of these slices, ensuring that the model's input does not exceed the limit in each instance.
For example, HiPool \cite{li2023hipool} employs Transformers for sentence modeling and then uses Graph Convolutional Neural Networks for document information modeling.
HiStruct+ \cite{ruan2022histruct} encodes the hierarchical structure information of the document and infuses it into the hierarchical attention model.
Due to the full-rank attention mechanism in transformer models leading to quadratic computational complexity, efficient transformers~\cite{ beltagy2020longformer, zaheer2020big, choromanski2021rethinking, zhang2023adaptive} aim to use sparse attention or low-rank methods to reduce the complexity and minimize context fragmentation caused by segmentation.
For instance, to reduce computational complexity from \( O(n^2) \) to \( O(n) \), Longformer~\cite{beltagy2020longformer} employs a mix of local attention (through a sliding window) and global attention on certain special tokens. 
Similarly, BigBird~\cite{zaheer2020big} incorporates both these attention mechanisms and introduces an additional random attention strategy.
Both models have expanded their input limits to 4096 tokens.
However, they do not perform well on documents of all lengths.

Prior research~\cite{li2023hipool} has noted that document lengths differ among datasets, and model performance can be inconsistent across corpora with varying lengths.
Studies~\cite{dai2022revisiting} have also shown that segmenting documents inevitably leads to issues of context fragmentation.
However, no previous work has centered on the aforementioned two inherent issues of long document models: context fragmentation and generalizability across varying text lengths.
In this study, we propose a novel approach Length-Aware Multi-Kernel Transformer (\textit{LAMKIT}). 
By using multi-kernel encoding (MK), LAMKIT obtains multi-perspective context representations to mitigate the context fragmentation issue caused by using a unique chunk size.
LAMKIT also enhances model robustness for documents of varying lengths through its Length-Aware Vectorization (LaV) module. This LaV module encodes length information hierarchically, using segment position embedding at the segment level and length vectors from the MK outputs at the document level.

\section{Conclusion}

In this study, we posit that for long document classification tasks, the length of the text might be a pivotal determinant for model performance. 
Our exploratory experiments demonstrate that the current state-of-the-art models display inconsistent results across samples of differing lengths, suggesting their lack of robustness and affirming our hypothesis. 

To address this issue and the inherent problem of context fragmentation in long-text models, we propose Length-Aware Multi-Kernel Transformer.
Through extensive experiments, LAMKIT consistently outperforms all baseline models across five standard long document classification benchmarks. 
Moreover, we follow our exploratory experiments to examine model robustness over varying document lengths. 
We also conduct ablation studies on two modules. 
The results show that LAMKIT exhibits better robustness and stability across different lengths.

Additionally, the case study on ChatGPT~\citep{openai2022chatgpt} reveals that LLMs still underperform discriminative models on long document classification tasks, suggesting that the paradigm of solving classification problems through generation still needs to be enhanced.

\section*{Limitations}

LAMKIT has a flexibility to be applicable on other tasks by changing its prediction layer, while we experiment it on the text classification task.
\citeauthor{dong2023survey} demonstrated the importance of long document modeling in other NLP scenarios.
We plan to explore this direction for a more comprehensive understanding on long document modeling.

\section{Acknowledgement}

This work was partially supported by the National Science Foundation by award number CNS-2318210.  
We thank anonymous reviewers for their insightful feedback.


\begin{thebibliography}{54}
\expandafter\ifx\csname natexlab\endcsname\relax\def\natexlab#1{#1}\fi

\bibitem[{Alsentzer et~al.(2019)Alsentzer, Murphy, Boag, Weng, Jindi, Naumann, and McDermott}]{alsentzer2019publicly}
Emily Alsentzer, John Murphy, William Boag, Wei-Hung Weng, Di~Jindi, Tristan Naumann, and Matthew McDermott. 2019.
\newblock \href {https://doi.org/10.18653/v1/W19-1909} {Publicly available clinical {BERT} embeddings}.
\newblock In \emph{Proceedings of the 2nd Clinical Natural Language Processing Workshop}, pages 72--78, Minneapolis, Minnesota, USA. Association for Computational Linguistics.

\bibitem[{Beltagy et~al.(2020)Beltagy, Peters, and Cohan}]{beltagy2020longformer}
Iz~Beltagy, Matthew~E. Peters, and Arman Cohan. 2020.
\newblock \href {http://arxiv.org/abs/2004.05150} {Longformer: The long-document transformer}.
\newblock \emph{arXiv preprint arXiv:2004.05150}.

\bibitem[{Chalkidis(2023)}]{chalkidis2023chatgpt}
Ilias Chalkidis. 2023.
\newblock \href {https://doi.org/10.2139/ssrn.4385460} {{ChatGPT May Pass the Bar Exam Soon, but Has a Long Way to Go for the LexGLUE Benchmark}}.
\newblock \emph{SSRN Electronic Journal}.

\bibitem[{Chalkidis et~al.(2019)Chalkidis, Androutsopoulos, and Aletras}]{chalkidis2019neural}
Ilias Chalkidis, Ion Androutsopoulos, and Nikolaos Aletras. 2019.
\newblock \href {https://doi.org/10.18653/v1/P19-1424} {Neural legal judgment prediction in {E}nglish}.
\newblock In \emph{Proceedings of the 57th Annual Meeting of the Association for Computational Linguistics}, pages 4317--4323, Florence, Italy. Association for Computational Linguistics.

\bibitem[{Chalkidis et~al.(2020)Chalkidis, Fergadiotis, Malakasiotis, Aletras, and Androutsopoulos}]{chalkidis2020legal}
Ilias Chalkidis, Manos Fergadiotis, Prodromos Malakasiotis, Nikolaos Aletras, and Ion Androutsopoulos. 2020.
\newblock \href {https://doi.org/10.18653/v1/2020.findings-emnlp.261} {{LEGAL}-{BERT}: The muppets straight out of law school}.
\newblock In \emph{Findings of the Association for Computational Linguistics: EMNLP 2020}, pages 2898--2904, Online. Association for Computational Linguistics.

\bibitem[{Chalkidis et~al.(2021)Chalkidis, Fergadiotis, Tsarapatsanis, Aletras, Androutsopoulos, and Malakasiotis}]{chalkidis2021paragraph}
Ilias Chalkidis, Manos Fergadiotis, Dimitrios Tsarapatsanis, Nikolaos Aletras, Ion Androutsopoulos, and Prodromos Malakasiotis. 2021.
\newblock \href {https://doi.org/10.18653/v1/2021.naacl-main.22} {Paragraph-level rationale extraction through regularization: A case study on {E}uropean court of human rights cases}.
\newblock In \emph{Proceedings of the 2021 Conference of the North American Chapter of the Association for Computational Linguistics: Human Language Technologies}, pages 226--241, Online. Association for Computational Linguistics.

\bibitem[{Chalkidis et~al.(2023)Chalkidis, Garneau, Goanta, Katz, and S{\o}gaard}]{chalkidis2023lexfiles}
Ilias Chalkidis, Nicolas Garneau, Catalina Goanta, Daniel Katz, and Anders S{\o}gaard. 2023.
\newblock \href {https://doi.org/10.18653/v1/2023.acl-long.865} {{L}e{XF}iles and {L}egal{LAMA}: Facilitating {E}nglish multinational legal language model development}.
\newblock In \emph{Proceedings of the 61st Annual Meeting of the Association for Computational Linguistics (Volume 1: Long Papers)}, pages 15513--15535, Toronto, Canada. Association for Computational Linguistics.

\bibitem[{Chalkidis et~al.(2022)Chalkidis, Jana, Hartung, Bommarito, Androutsopoulos, Katz, and Aletras}]{chalkidis2022lexglue}
Ilias Chalkidis, Abhik Jana, Dirk Hartung, Michael Bommarito, Ion Androutsopoulos, Daniel Katz, and Nikolaos Aletras. 2022.
\newblock \href {https://doi.org/10.18653/v1/2022.acl-long.297} {{L}ex{GLUE}: A benchmark dataset for legal language understanding in {E}nglish}.
\newblock In \emph{Proceedings of the 60th Annual Meeting of the Association for Computational Linguistics (Volume 1: Long Papers)}, pages 4310--4330, Dublin, Ireland. Association for Computational Linguistics.

\bibitem[{Chen et~al.(2023{\natexlab{a}})Chen, Chen, Huang, and Zhou}]{chen2023need}
Jiuhai Chen, Lichang Chen, Heng Huang, and Tianyi Zhou. 2023{\natexlab{a}}.
\newblock \href {http://arxiv.org/abs/2304.03262} {When do you need chain-of-thought prompting for chatgpt?}
\newblock \emph{arXiv preprint arXiv:2304.03262}.

\bibitem[{Chen et~al.(2023{\natexlab{b}})Chen, Chen, Zhu, and Zhou}]{chen2023many}
Jiuhai Chen, Lichang Chen, Chen Zhu, and Tianyi Zhou. 2023{\natexlab{b}}.
\newblock \href {https://doi.org/10.18653/v1/2023.findings-emnlp.745} {How many demonstrations do you need for in-context learning?}
\newblock In \emph{Findings of the Association for Computational Linguistics: EMNLP 2023}, pages 11149--11159, Singapore. Association for Computational Linguistics.

\bibitem[{Choromanski et~al.(2021)Choromanski, Likhosherstov, Dohan, Song, Gane, Sarlos, Hawkins, Davis, Mohiuddin, Kaiser, Belanger, Colwell, and Weller}]{choromanski2021rethinking}
Krzysztof~Marcin Choromanski, Valerii Likhosherstov, David Dohan, Xingyou Song, Andreea Gane, Tamas Sarlos, Peter Hawkins, Jared~Quincy Davis, Afroz Mohiuddin, Lukasz Kaiser, David~Benjamin Belanger, Lucy~J Colwell, and Adrian Weller. 2021.
\newblock \href {https://openreview.net/forum?id=Ua6zuk0WRH} {Rethinking attention with performers}.
\newblock In \emph{International Conference on Learning Representations}, Vienna, Austria.

\bibitem[{Dai et~al.(2022)Dai, Chalkidis, Darkner, and Elliott}]{dai2022revisiting}
Xiang Dai, Ilias Chalkidis, Sune Darkner, and Desmond Elliott. 2022.
\newblock \href {https://doi.org/10.18653/v1/2022.findings-emnlp.534} {Revisiting transformer-based models for long document classification}.
\newblock In \emph{Findings of the Association for Computational Linguistics: EMNLP 2022}, pages 7212--7230, Abu Dhabi, United Arab Emirates. Association for Computational Linguistics.

\bibitem[{Dassi and Kwate(2021)}]{dassi2021legal}
Loic~Kwate Dassi and Loic Kwate. 2021.
\newblock \href {https://neurips.cc/virtual/2021/48411} {Legal-bigbird: An adapted long-range transformer for legal documents}.
\newblock In \emph{Proceedings of the 35th International Conference on Neural Information Processing Systems, Black in AI Workshop}. Curran Associates, Inc.

\bibitem[{Devlin et~al.(2019)Devlin, Chang, Lee, and Toutanova}]{devlin2019bert}
Jacob Devlin, Ming-Wei Chang, Kenton Lee, and Kristina Toutanova. 2019.
\newblock \href {https://doi.org/10.18653/v1/N19-1423} {{BERT}: Pre-training of deep bidirectional transformers for language understanding}.
\newblock In \emph{Proceedings of the 2019 Conference of the North {A}merican Chapter of the Association for Computational Linguistics: Human Language Technologies, Volume 1 (Long and Short Papers)}, pages 4171--4186, Minneapolis, Minnesota. Association for Computational Linguistics.

\bibitem[{Ding et~al.(2020)Ding, Zhou, Yang, and Tang}]{ding2020cogltx}
Ming Ding, Chang Zhou, Hongxia Yang, and Jie Tang. 2020.
\newblock \href {https://proceedings.neurips.cc/paper_files/paper/2020/file/96671501524948bc3937b4b30d0e57b9-Paper.pdf} {Cogltx: Applying bert to long texts}.
\newblock In \emph{Advances in Neural Information Processing Systems}, volume~33, pages 12792--12804, Vancouver, British Columbia, Canada. Curran Associates, Inc.

\bibitem[{Dong et~al.(2023)Dong, Tang, Li, and Zhao}]{dong2023survey}
Zican Dong, Tianyi Tang, Lunyi Li, and Wayne~Xin Zhao. 2023.
\newblock \href {http://arxiv.org/abs/2302.14502} {A survey on long text modeling with transformers}.
\newblock \emph{arXiv preprint arXiv:2302.14502}.

\bibitem[{Guo et~al.(2022)Guo, Ainslie, Uthus, Ontanon, Ni, Sung, and Yang}]{guo2022longt5}
Mandy Guo, Joshua Ainslie, David Uthus, Santiago Ontanon, Jianmo Ni, Yun-Hsuan Sung, and Yinfei Yang. 2022.
\newblock \href {https://doi.org/10.18653/v1/2022.findings-naacl.55} {{L}ong{T}5: {E}fficient text-to-text transformer for long sequences}.
\newblock In \emph{Findings of the Association for Computational Linguistics: NAACL 2022}, pages 724--736, Seattle, United States. Association for Computational Linguistics.

\bibitem[{Han et~al.(2024)Han, Liu, Huang, and Borsari}]{han2024chainofinteraction}
Guangzeng Han, Weisi Liu, Xiaolei Huang, and Brian Borsari. 2024.
\newblock \href {http://arxiv.org/abs/2403.13786} {Chain-of-interaction: Enhancing large language models for psychiatric behavior understanding by dyadic contexts}.
\newblock \emph{arXiv preprint arXiv:2403.13786}.

\bibitem[{He et~al.(2021)He, Liu, Gao, and Chen}]{he2021deberta}
Pengcheng He, Xiaodong Liu, Jianfeng Gao, and Weizhu Chen. 2021.
\newblock \href {https://openreview.net/forum?id=XPZIaotutsD} {Deberta: Decoding enhanced bert with disentangled attention}.
\newblock In \emph{International Conference on Learning Representations}.

\bibitem[{Jin and Wang(2023)}]{jin2023understand}
Xin Jin and Yuchen Wang. 2023.
\newblock \href {http://arxiv.org/abs/2303.12135} {Understand legal documents with contextualized large language models}.
\newblock \emph{arXiv preprint arXiv:2303.12135}.

\bibitem[{Johnson et~al.(2016)Johnson, Pollard, Shen, Lehman, Feng, Ghassemi, Moody, Szolovits, Celi, and Mark}]{johnson2016mimic}
Alistair~EW Johnson, Tom~J Pollard, Lu~Shen, Li{-}wei~H Lehman, Mengling Feng, Mohammad Ghassemi, Benjamin Moody, Peter Szolovits, Leo~Anthony Celi, and Roger~G Mark. 2016.
\newblock \href {https://doi.org/10.1038/sdata.2016.35} {Mimic-iii, a freely accessible critical care database}.
\newblock \emph{Scientific Data}, 3:160035.

\bibitem[{Kim(2014)}]{kim2014convolutional}
Yoon Kim. 2014.
\newblock \href {https://doi.org/10.3115/v1/D14-1181} {Convolutional neural networks for sentence classification}.
\newblock In \emph{Proceedings of the 2014 Conference on Empirical Methods in Natural Language Processing ({EMNLP})}, pages 1746--1751, Doha, Qatar. Association for Computational Linguistics.

\bibitem[{Lewis et~al.(2020)Lewis, Ott, Du, and Stoyanov}]{lewis2020pretrained}
Patrick Lewis, Myle Ott, Jingfei Du, and Veselin Stoyanov. 2020.
\newblock \href {https://doi.org/10.18653/v1/2020.clinicalnlp-1.17} {Pretrained language models for biomedical and clinical tasks: Understanding and extending the state-of-the-art}.
\newblock In \emph{Proceedings of the 3rd Clinical Natural Language Processing Workshop}, pages 146--157, Online. Association for Computational Linguistics.

\bibitem[{Li et~al.(2023{\natexlab{a}})Li, Feng, Radev, and Ying}]{li2023hipool}
Irene Li, Aosong Feng, Dragomir Radev, and Rex Ying. 2023{\natexlab{a}}.
\newblock \href {https://doi.org/10.18653/v1/2023.acl-short.16} {{H}i{P}ool: Modeling long documents using graph neural networks}.
\newblock In \emph{Proceedings of the 61st Annual Meeting of the Association for Computational Linguistics (Volume 2: Short Papers)}, pages 161--171, Toronto, Canada. Association for Computational Linguistics.

\bibitem[{Li et~al.(2023{\natexlab{b}})Li, Li, Luo, Xie, Lee, Zhao, Wang, and Li}]{li2023recurrent}
Xianming Li, Zongxi Li, Xiaotian Luo, Haoran Xie, Xing Lee, Yingbin Zhao, Fu~Lee Wang, and Qing Li. 2023{\natexlab{b}}.
\newblock \href {https://doi.org/10.18653/v1/2023.findings-acl.188} {Recurrent attention networks for long-text modeling}.
\newblock In \emph{Findings of the Association for Computational Linguistics: ACL 2023}, pages 3006--3019, Toronto, Canada. Association for Computational Linguistics.

\bibitem[{Li et~al.(2023{\natexlab{c}})Li, Wehbe, Ahmad, Wang, and Luo}]{li2023comparative}
Yikuan Li, Ramsey~M Wehbe, Faraz~S Ahmad, Hanyin Wang, and Yuan Luo. 2023{\natexlab{c}}.
\newblock \href {https://doi.org/10.1093/jamia/ocac225} {{A comparative study of pretrained language models for long clinical text}}.
\newblock \emph{Journal of the American Medical Informatics Association}, 30(2):340--347.

\bibitem[{Liu et~al.(2019)Liu, Ott, Goyal, Du, Joshi, Chen, Levy, Lewis, Zettlemoyer, and Stoyanov}]{liu2019roberta}
Yinhan Liu, Myle Ott, Naman Goyal, Jingfei Du, Mandar Joshi, Danqi Chen, Omer Levy, Mike Lewis, Luke Zettlemoyer, and Veselin Stoyanov. 2019.
\newblock \href {http://arxiv.org/abs/1907.11692} {Roberta: A robustly optimized bert pretraining approach}.
\newblock \emph{arXiv preprint arXiv:1907.11692}.

\bibitem[{Loshchilov and Hutter(2019)}]{loshchilov2019decoupled}
Ilya Loshchilov and Frank Hutter. 2019.
\newblock \href {https://openreview.net/forum?id=Bkg6RiCqY7} {Decoupled weight decay regularization}.
\newblock In \emph{International Conference on Learning Representations}.

\bibitem[{Lou et~al.(2023)Lou, Zhang, and Yin}]{lou2023prompt}
Renze Lou, Kai Zhang, and Wenpeng Yin. 2023.
\newblock \href {http://arxiv.org/abs/2303.10475} {Is prompt all you need? no. a comprehensive and broader view of instruction learning}.
\newblock \emph{arXiv preprint arXiv:2303.10475}.

\bibitem[{Ma et~al.(2021)Ma, Zhang, Lou, Wang, and Vosoughi}]{ma2021contributions}
Weicheng Ma, Kai Zhang, Renze Lou, Lili Wang, and Soroush Vosoughi. 2021.
\newblock \href {https://doi.org/10.18653/v1/2021.acl-long.152} {Contributions of transformer attention heads in multi- and cross-lingual tasks}.
\newblock In \emph{Proceedings of the 59th Annual Meeting of the Association for Computational Linguistics and the 11th International Joint Conference on Natural Language Processing (Volume 1: Long Papers)}, pages 1956--1966, Online. Association for Computational Linguistics.

\bibitem[{Mullenbach et~al.(2018)Mullenbach, Wiegreffe, Duke, Sun, and Eisenstein}]{mullenbach2018explainable}
James Mullenbach, Sarah Wiegreffe, Jon Duke, Jimeng Sun, and Jacob Eisenstein. 2018.
\newblock \href {https://doi.org/10.18653/v1/N18-1100} {Explainable prediction of medical codes from clinical text}.
\newblock In \emph{Proceedings of the 2018 Conference of the North {A}merican Chapter of the Association for Computational Linguistics: Human Language Technologies, Volume 1 (Long Papers)}, pages 1101--1111, New Orleans, Louisiana. Association for Computational Linguistics.

\bibitem[{Nerella et~al.(2023)Nerella, Bandyopadhyay, Zhang, Contreras, Siegel, Bumin, Silva, Sena, Shickel, Bihorac et~al.}]{nerella2023transformers}
Subhash Nerella, Sabyasachi Bandyopadhyay, Jiaqing Zhang, Miguel Contreras, Scott Siegel, Aysegul Bumin, Brandon Silva, Jessica Sena, Benjamin Shickel, Azra Bihorac, et~al. 2023.
\newblock Transformers in healthcare: A survey.
\newblock \emph{arXiv preprint arXiv:2307.00067}.

\bibitem[{OpenAI(2022)}]{openai2022chatgpt}
OpenAI. 2022.
\newblock Chatgpt: Optimizing language models for dialogue.
\newblock \url{https://openai.com/blog/chatgpt/}.
\newblock Accessed: 2023-07-24.

\bibitem[{Pan et~al.(2024)Pan, Luo, Li, and Liu}]{pan2024chainofaction}
Zhenyu Pan, Haozheng Luo, Manling Li, and Han Liu. 2024.
\newblock \href {http://arxiv.org/abs/2403.17359} {Chain-of-action: Faithful and multimodal question answering through large language models}.
\newblock \emph{arXiv preprint arXiv:2403.17359}.

\bibitem[{Paszke et~al.(2019)Paszke, Gross, Massa, Lerer, Bradbury, Chanan, Killeen, Lin, Gimelshein, Antiga, Desmaison, Kopf, Yang, DeVito, Raison, Tejani, Chilamkurthy, Steiner, Fang, Bai, and Chintala}]{paszke2019pytorch}
Adam Paszke, Sam Gross, Francisco Massa, Adam Lerer, James Bradbury, Gregory Chanan, Trevor Killeen, Zeming Lin, Natalia Gimelshein, Luca Antiga, Alban Desmaison, Andreas Kopf, Edward Yang, Zachary DeVito, Martin Raison, Alykhan Tejani, Sasank Chilamkurthy, Benoit Steiner, Lu~Fang, Junjie Bai, and Soumith Chintala. 2019.
\newblock \href {https://proceedings.neurips.cc/paper_files/paper/2019/file/bdbca288fee7f92f2bfa9f7012727740-Paper.pdf} {Pytorch: An imperative style, high-performance deep learning library}.
\newblock In \emph{Advances in Neural Information Processing Systems}, volume~32, pages 8024--8035. Curran Associates, Inc.

\bibitem[{Qin et~al.(2023)Qin, Feng, and Van~Durme}]{qin2023nlp}
Guanghui Qin, Yukun Feng, and Benjamin Van~Durme. 2023.
\newblock \href {https://aclanthology.org/2023.eacl-main.273} {The {NLP} task effectiveness of long-range transformers}.
\newblock In \emph{Proceedings of the 17th Conference of the European Chapter of the Association for Computational Linguistics}, pages 3774--3790, Dubrovnik, Croatia. Association for Computational Linguistics.

\bibitem[{Ruan et~al.(2022)Ruan, Ostendorff, and Rehm}]{ruan2022histruct}
Qian Ruan, Malte Ostendorff, and Georg Rehm. 2022.
\newblock \href {https://doi.org/10.18653/v1/2022.findings-acl.102} {{H}i{S}truct+: Improving extractive text summarization with hierarchical structure information}.
\newblock In \emph{Findings of the Association for Computational Linguistics: ACL 2022}, pages 1292--1308, Dublin, Ireland. Association for Computational Linguistics.

\bibitem[{Rule et~al.(2021)Rule, Bedrick, Chiang, and Hribar}]{rule2021length}
Adam Rule, Steven Bedrick, Michael~F. Chiang, and Michelle~R. Hribar. 2021.
\newblock \href {https://doi.org/10.1001/jamanetworkopen.2021.15334} {{Length and Redundancy of Outpatient Progress Notes Across a Decade at an Academic Medical Center}}.
\newblock \emph{JAMA Network Open}, 4(7):e2115334--e2115334.

\bibitem[{Song et~al.(2023)Song, Wu, Zhang, Peng, Dang, Pan, and Wu}]{song23ZeroPrompt}
Xingchen Song, Di~Wu, Binbin Zhang, Zhendong Peng, Bo~Dang, Fuping Pan, and Zhiyong Wu. 2023.
\newblock \href {https://doi.org/10.21437/Interspeech.2023-1497} {Zeroprompt: Streaming acoustic encoders are zero-shot masked lms}.
\newblock In \emph{Proc. INTERSPEECH 2023}, pages 1648--1652.

\bibitem[{Spaeth et~al.(2020)Spaeth, Epstein, Martin, Segal, Ruger, and Benesh}]{spaeth2020Supreme}
Harold~J. Spaeth, Lee Epstein, Andrew~D. Martin, Jeffrey~A. Segal, Theodore~J. Ruger, and Sara~C. Benesh. 2020.
\newblock {Supreme Court Database, Version 2020 Release 01}.
\newblock \url{http://Supremecourtdatabase.org}.
\newblock Accessed: [2021-01-01].

\bibitem[{Srivastava et~al.(2023)Srivastava, Huang, Fan, and Yao}]{srivastava2023instance}
Saurabh Srivastava, Chengyue Huang, Weiguo Fan, and Ziyu Yao. 2023.
\newblock Instance needs more care: Rewriting prompts for instances yields better zero-shot performance.
\newblock \emph{arXiv preprint arXiv:2310.02107}.

\bibitem[{Stubbs et~al.(2019)Stubbs, Filannino, Soysal, Henry, and Uzuner}]{stubbs2019cohort}
Amber Stubbs, Michele Filannino, Ergin Soysal, Samuel Henry, and {\"O}zlem Uzuner. 2019.
\newblock \href {https://doi.org/10.1093/jamia/ocz163} {{Cohort selection for clinical trials: n2c2 2018 shared task track 1}}.
\newblock \emph{Journal of the American Medical Informatics Association}, 26(11):1163--1171.

\bibitem[{Sun et~al.(2024)Sun, Ahmed, Ma, Liu, Kabela, Pang, and Kalinli}]{sun2024Contextual}
Chuanneng Sun, Zeeshan Ahmed, Yingyi Ma, Zhe Liu, Lucas Kabela, Yutong Pang, and Ozlem Kalinli. 2024.
\newblock \href {https://doi.org/10.1109/ICASSP48485.2024.10445918} {Contextual biasing of named-entities with large language models}.
\newblock In \emph{ICASSP 2024 - 2024 IEEE International Conference on Acoustics, Speech and Signal Processing (ICASSP)}, pages 10151--10155.

\bibitem[{Vaswani et~al.(2017)Vaswani, Shazeer, Parmar, Uszkoreit, Jones, Gomez, Kaiser, and Polosukhin}]{vaswani2017attention}
Ashish Vaswani, Noam Shazeer, Niki Parmar, Jakob Uszkoreit, Llion Jones, Aidan~N Gomez, \L~ukasz Kaiser, and Illia Polosukhin. 2017.
\newblock \href {https://proceedings.neurips.cc/paper_files/paper/2017/file/3f5ee243547dee91fbd053c1c4a845aa-Paper.pdf} {Attention is all you need}.
\newblock In \emph{Advances in Neural Information Processing Systems}, volume~30, Long Beach, California, United States. Curran Associates, Inc.

\bibitem[{Vu et~al.(2021)Vu, Nguyen, and Nguyen}]{vu2021label}
Thanh Vu, Dat~Quoc Nguyen, and Anthony Nguyen. 2021.
\newblock \href {https://doi.org/10.24963/ijcai.2020/461} {A label attention model for icd coding from clinical text}.
\newblock In \emph{Proceedings of the Twenty-Ninth International Joint Conference on Artificial Intelligence}, IJCAI'20, pages 3335--3341. International Joint Conferences on Artificial Intelligence Organization.
\newblock Main track.

\bibitem[{Wang et~al.(2023)Wang, Wei, Schuurmans, Le, Chi, Narang, Chowdhery, and Zhou}]{wang2023selfconsistency}
Xuezhi Wang, Jason Wei, Dale Schuurmans, Quoc Le, Ed~Chi, Sharan Narang, Aakanksha Chowdhery, and Denny Zhou. 2023.
\newblock \href {http://arxiv.org/abs/2203.11171} {Self-consistency improves chain of thought reasoning in language models}.
\newblock \emph{arXiv preprint arXiv:2203.11171}.

\bibitem[{Wei et~al.(2022)Wei, Wang, Schuurmans, Bosma, ichter, Xia, Chi, Le, and Zhou}]{Wei2022Chain}
Jason Wei, Xuezhi Wang, Dale Schuurmans, Maarten Bosma, brian ichter, Fei Xia, Ed~Chi, Quoc~V Le, and Denny Zhou. 2022.
\newblock \href {https://proceedings.neurips.cc/paper_files/paper/2022/file/9d5609613524ecf4f15af0f7b31abca4-Paper-Conference.pdf} {Chain-of-thought prompting elicits reasoning in large language models}.
\newblock In \emph{Advances in Neural Information Processing Systems}, volume~35, pages 24824--24837. Curran Associates, Inc.

\bibitem[{Wolf et~al.(2020)Wolf, Debut, Sanh, Chaumond, Delangue, Moi, Cistac, Rault, Louf, Funtowicz, Davison, Shleifer, von Platen, Ma, Jernite, Plu, Xu, Le~Scao, Gugger, Drame, Lhoest, and Rush}]{wolf2020transformers}
Thomas Wolf, Lysandre Debut, Victor Sanh, Julien Chaumond, Clement Delangue, Anthony Moi, Pierric Cistac, Tim Rault, Remi Louf, Morgan Funtowicz, Joe Davison, Sam Shleifer, Patrick von Platen, Clara Ma, Yacine Jernite, Julien Plu, Canwen Xu, Teven Le~Scao, Sylvain Gugger, Mariama Drame, Quentin Lhoest, and Alexander Rush. 2020.
\newblock \href {https://doi.org/10.18653/v1/2020.emnlp-demos.6} {Transformers: State-of-the-art natural language processing}.
\newblock In \emph{Proceedings of the 2020 Conference on Empirical Methods in Natural Language Processing: System Demonstrations}, pages 38--45, Online. Association for Computational Linguistics.

\bibitem[{Wu et~al.(2021)Wu, Wu, Qi, and Huang}]{wu2021hi}
Chuhan Wu, Fangzhao Wu, Tao Qi, and Yongfeng Huang. 2021.
\newblock \href {https://doi.org/10.18653/v1/2021.acl-short.107} {Hi-transformer: Hierarchical interactive transformer for efficient and effective long document modeling}.
\newblock In \emph{Proceedings of the 59th Annual Meeting of the Association for Computational Linguistics and the 11th International Joint Conference on Natural Language Processing (Volume 2: Short Papers)}, pages 848--853, Online. Association for Computational Linguistics.

\bibitem[{Xiong et~al.(2024)Xiong, Payani, Kompella, and Fekri}]{xiong2024large}
Siheng Xiong, Ali Payani, Ramana Kompella, and Faramarz Fekri. 2024.
\newblock Large language models can learn temporal reasoning.
\newblock \emph{arXiv preprint arXiv:2401.06853}.

\bibitem[{Zaheer et~al.(2020)Zaheer, Guruganesh, Dubey, Ainslie, Alberti, Ontanon, Pham, Ravula, Wang, Yang, and Ahmed}]{zaheer2020big}
Manzil Zaheer, Guru Guruganesh, Avinava Dubey, Joshua Ainslie, Chris Alberti, Santiago Ontanon, Philip Pham, Anirudh Ravula, Qifan Wang, Li~Yang, and Amr Ahmed. 2020.
\newblock \href {https://proceedings.neurips.cc/paper/2020/file/c8512d142a2d849725f31a9a7a361ab9-Paper.pdf} {Big bird: Transformers for longer sequences}.
\newblock In \emph{Proceedings of the 34th International Conference on Neural Information Processing Systems}, volume~33, pages 17283--17297, Red Hook, NY, USA. Curran Associates Inc.

\bibitem[{Zhang et~al.(2023)Zhang, Lv, and Yang}]{zhang2023adaptive}
Xuanyu Zhang, Zhepeng Lv, and Qing Yang. 2023.
\newblock \href {https://doi.org/10.18653/v1/2023.findings-acl.546} {Adaptive attention for sparse-based long-sequence transformer}.
\newblock In \emph{Findings of the Association for Computational Linguistics: ACL 2023}, pages 8602--8610, Toronto, Canada. Association for Computational Linguistics.

\bibitem[{Zhang et~al.(2024)Zhang, Gui, Zhu, Hao, and Sun}]{zhang2024unlocking}
Ye~Zhang, Kailin Gui, Mengran Zhu, Yong Hao, and Haozhan Sun. 2024.
\newblock Unlocking personalized anime recommendations: Langchain and llm at the forefront.
\newblock \emph{Journal of Industrial Engineering and Applied Science}, 2(2):46--53.

\bibitem[{Zhou et~al.(2022)Zhou, Geng, Shen, Long, and Jiang}]{zhou2022eventbert}
Yucheng Zhou, Xiubo Geng, Tao Shen, Guodong Long, and Daxin Jiang. 2022.
\newblock \href {https://doi.org/10.1145/3485447.3511928} {Eventbert: A pre-trained model for event correlation reasoning}.
\newblock In \emph{Proceedings of the ACM Web Conference 2022}, WWW '22, page 850–859, New York, NY, USA. Association for Computing Machinery.

\end{thebibliography}

\appendix
\section{Experimental Details}
\label{sec:implementation}

For all baseline models, we maintain the same model architecture and optimization parameters as described in their respective papers. 
For Longformer~\cite{beltagy2020longformer}, Bigbird~\cite{zaheer2020big}, and BERT\cite{devlin2019bert}, we fine-tune the pre-trained models obtained from huggingface transformers \cite{wolf2020transformers} library based on their given configurations and produce predictions.
For H-BERT\cite{dai2022revisiting}, we train using the code released by the authors and obtain our results.

For our proposed  \textit{LAMKIT} model. 
The kernel sizes are set to \{32, 64, 128\} in the ECTHR dataset and \{128, 256, 512\} in the other three datasets. 
The corresponding segment numbers are set to \{128, 64, 32\} and \{32, 16, 8\} to ensure that the input length of LAMKIT is 4096 tokens, the same as the other baselines.
The kernel stride is set by default to be equal to the kernel size.
To make the results reproducible, we set the random seed in training to 1.
For the MIMIC-III and Diabetes datasets, we employ pretrained Roberta-PM-M3-base \cite{lewis2020pretrained} as our multi-kernel encoder. 
For SCOTUS and ECtHR, we opt for pretrained Legal-BERT-base \cite{chalkidis2020legal}. 
Both encoders have 12 layers, 12 attention heads, and hidden states of 768 dimensions. 
Additionally, we set a Transformer~\cite{vaswani2017attention} encoder with 1 layer, 12 attention heads, and 768-dimensional hidden states as the length encoder, and another with 2 layers, 12 attention heads, and 768-dimensional hidden states as the document encoder. 
The dropout between the two linear layers of the classifier is set at 0.1.
Due to our limited computational resources, we empirically set the learning rate and tried two batch sizes: 32 and 16. 
Each experiment is set with a maximum of 20 training epochs and an early stopping patience of 3. 
We utilize the AdamW \cite{loshchilov2019decoupled} optimizer, with a weight decay of 0.01. 
To expedite model convergence, we make use of 16-bit float point numbers (half-precision). 
Finally, we select the best-performing model based on F1-micro on the validation set. 
The chosen hyperparameters for the model are presented in table~\ref{tab:experiments}.

\begin{table}[ht]
\centering
\resizebox{0.47\textwidth}{!}{%
\begin{tabular}{|l|c|c|c|c|c|}
\hline
  Dataset& Learing Rate & Batch Size & \multicolumn{3}{c|}{Kernel Size}  \\ \hline
MIMIC  & 3.5e-5 & 16 & 128 & 256 & 512 \\
ECtHR  & 1.0e-5 & 32 & 32 & 64 & 128 \\
SCOTUS  & 3.5e-5 & 16 & 128 & 256 & 512  \\
Diabetes  & 2.5e-5 & 16 & 128 & 256 & 512 \\ \hline
\end{tabular}%
}
\caption{Chosen hyperparameters for LAMKIT.}
\label{tab:experiments}
\end{table}

All experiments are conducted on a device equipped with an NVIDIA 3090 GPU with 24GB memory, running the Ubuntu system, and utilizing the PyTorch \cite{paszke2019pytorch} framework.

\newpage
\section{Prompt Template of Case Study}
\label{sec:appendix}

\begin{figure*}[ht]
\centering
\includegraphics[width=\textwidth]{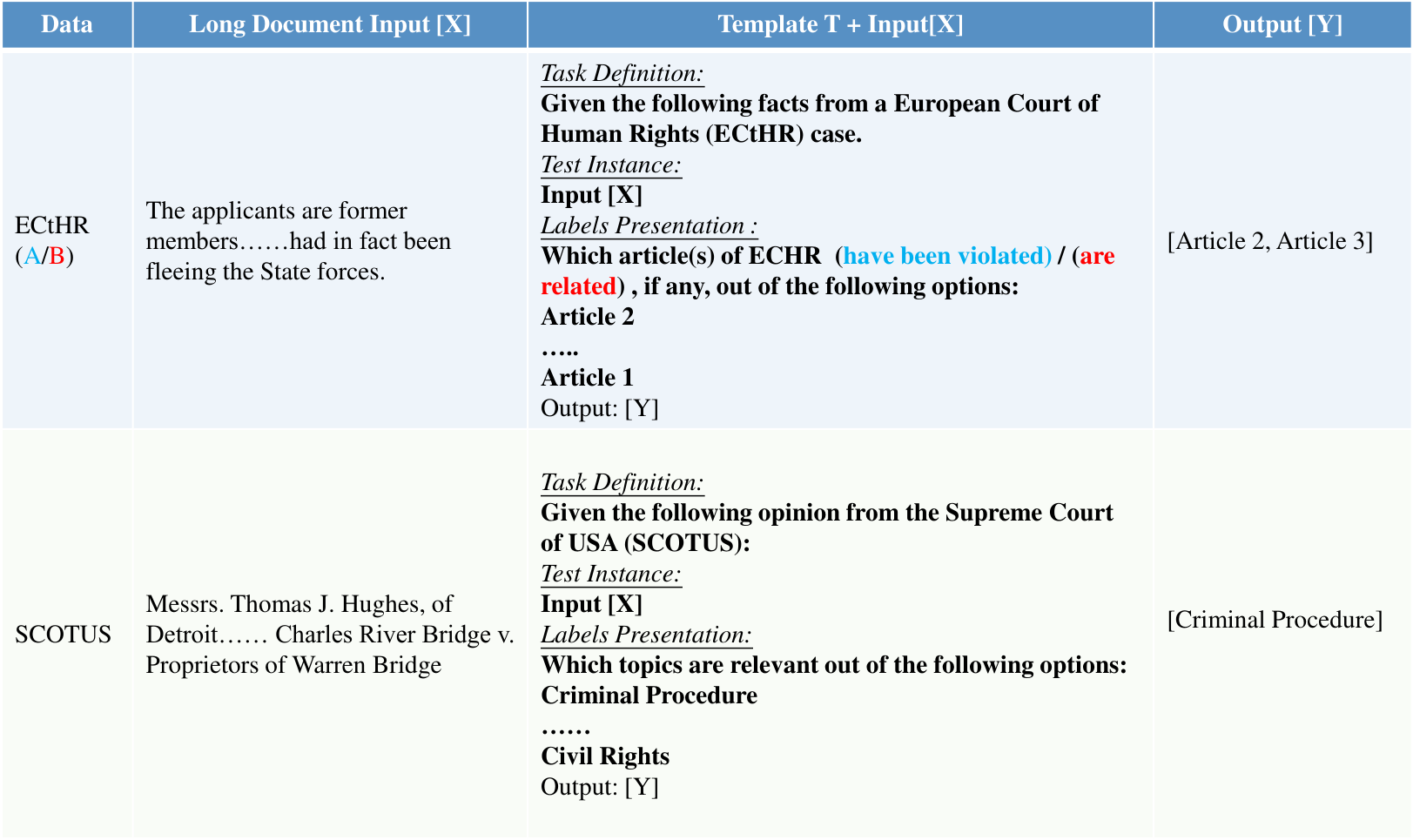}
\caption{The best performing zero-shot template of the legal data.}
\label{fig:template}

\end{figure*}

For ChatGPT~\cite{openai2022chatgpt}, we set the temperature to 0, and the Top P sampling value to 1. The prompt template is shown in Figure~\ref{fig:template}.

\end{document}